\documentclass{ecai}
\usepackage{latexsym}
\usepackage{amssymb}
\usepackage{amsmath}
\usepackage{amsthm}
\usepackage{booktabs}
\usepackage{enumitem}
\usepackage{graphicx}
\usepackage{color}
\usepackage[bb=dsserif]{mathalpha}
\usepackage[ruled,linesnumbered]{algorithm2e}
\usepackage[table,xcdraw]{xcolor}
\usepackage{multirow}

\newcommand{\BibTeX}{B\kern-.05em{\sc i\kern-.025em b}\kern-.08em\TeX}

\begin{document}

%%%%%%%%%%%%%%%%%%%%%%%%%%%%%%%%%%%%%%%%%%%%%%%%%%%%%%%%%%%%%%%%%%%%%%%%

\begin{frontmatter}

%%% Use this command to specify your submission number.
%%% In doubleblind mode, it will be printed on the first page.

\paperid{123} 

%%% Use this command to specify the title of your paper.

\title{Stabilizing Data-Free Model Extraction}

%%% Use this combinations of commands to specify all authors of your 
%%% paper. Use \fnms{} and \snm{} to indicate everyone's first names 
%%% and surname. This will help the publisher with indexing the 
%%% proceedings. Please use a reasonable approximation in case your 
%%% name does not neatly split into "first names" and "surname".
%%% Specifying your ORCID digital identifier is optional. 
%%% Use the \thanks{} command to indicate one or more corresponding 
%%% authors and their email address(es). If so desired, you can specify
%%% author contributions using the \footnote{} command.

\author[A]{\fnms{Dat-Thinh}~\snm{Nguyen}\thanks{Corresponding Author. Email: dat-thinh.nguyen@ucdconnect.ie}}
\author[B]{\fnms{Kim-Hung}~\snm{Le}}
\author[A]{\fnms{Nhien-An}~\snm{Le-Khac}\thanks{Corresponding Author. Email: an.lekhac@ucd.ie}} 

\address[A]{School of Computer Science, University College Dublin, Dublin, Ireland}
\address[B]{University of Information Technology, Vietnam National University, Ho Chi Minh City, Vietnam}

%%% Use this environment to include an abstract of your paper.

\begin{abstract}
Model extraction is a severe threat to Machine Learning-as-a-Service systems, especially through data-free approaches, where dishonest users can replicate the functionality of a black-box target model without access to realistic data. Despite recent advancements, existing data-free model extraction methods suffer from the oscillating accuracy of the substitute model. This oscillation, which could be attributed to the constant shift in the generated data distribution during the attack, makes the attack impractical since the optimal substitute model cannot be determined without access to the target model's in-distribution data. Hence, we propose MetaDFME, a novel data-free model extraction method that employs meta-learning in the generator training to reduce the distribution shift, aiming to mitigate the substitute model's accuracy oscillation. In detail, we train our generator to iteratively capture the meta-representations of the synthetic data during the attack. These meta-representations can be adapted with a few steps to produce data that facilitates the substitute model to learn from the target model while reducing the effect of distribution shifts. Our experiments on popular baseline image datasets, MNIST, SVHN, CIFAR-10, and CIFAR-100, demonstrate that MetaDFME outperforms the current state-of-the-art data-free model extraction method while exhibiting a more stable substitute model's accuracy during the attack.
\end{abstract}

\end{frontmatter}

\section{Introduction}
With recent breakthroughs in machine learning and the accessibility of cloud computing, Machine Learning-as-a-Service (MLaaS) has become prominent in various applications, where users can access high-quality predictive models with minimal costs. To access the serviced models, most MLaaS platforms enable users to make queries to and receive the corresponding predictions from the model via an application programming interface (API). Unfortunately, existing research has shown that serving the model to the public introduces a critical security threat called model extraction \cite{kesarwani2018model, tramer2016stealing}. This is an exploratory adversarial attack \cite{chakraborty2018adversarial}, where attackers can replicate the functionality from the black-box target model to an attacker-owned substitute model. Thus, model extraction not only threatens the business competitiveness of the MLaaS platforms but also exposes the model to further adversarial attacks.

Recent research in model extraction has advanced to data-free methods, which enable adversaries to clone the black-box target model without requiring access to the target's in-distribution data or public datasets \cite{kariyappa2021maze, truong2021data, rosenthal2023disguide}. These methods train a generative neural network to synthesize data that maximize the prediction disagreement between the target and substitute models (or among different substitute models). Meanwhile, the substitute model learns to match the target model's behaviors on this generated data. With this data-free approach, model extraction has been demonstrated effective in extracting high-performing image classification models \cite{rosenthal2023disguide}, highlighting the practicality and potential security threat of data-free model extraction (DFME) in the wild.

\begin{figure}[t]
    \centering
    \includegraphics[width=1\linewidth]{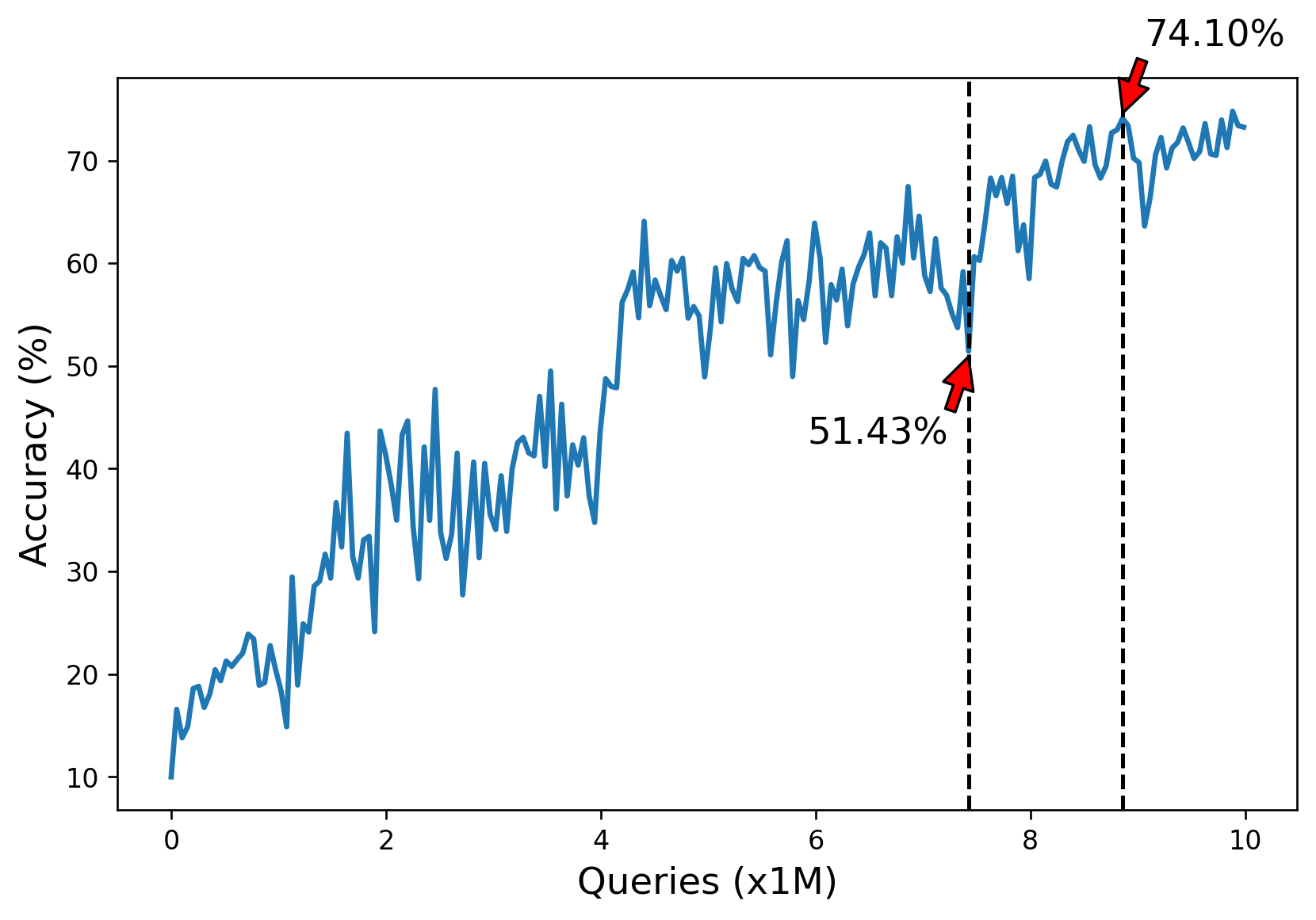}
    \caption{The accuracy of the substitute model from the state-of-the-art DFME method \cite{rosenthal2023disguide} on CIFAR-10 (see Section \ref{sec:experiment} for detailed experimental setup). Without access to the target's in-distribution data for periodic evaluation, the accuracy of the substitute model entirely relies on the random attack terminations. In this case, stopping the attack at 7.4M queries results in a sub-optimal substitute model with only 51.43\% accuracy, while stopping at 8.8M queries results in a better substitute model with 74.10\% accuracy.}
    \label{fig:problem_illustration}
\end{figure}

Despite these advancements, existing DFME methods suffer from the oscillating accuracy of the substitute model during the attack. This problem arises from the inherent training paradigm of these methods, which constantly alternates between generator and substitute model training. Under this paradigm, the generated data distribution is shifted every time the generator parameters are updated, making the substitute model learn different data distributions among attack iterations. Consequently, the accuracy of the substitute model on the target model's in-distribution data oscillates unpredictably during the attack. In data-free scenarios, this inconsistent performance is a non-trivial challenge because attackers cannot access the target's in-distribution data to periodically evaluate the substitute model and select the best checkpoint when the attack completes, as illustrated in Figure \ref{fig:problem_illustration}. To the best of our knowledge, no existing works in DFME have attempted to address this problem; hence, stabilizing DFME attacks is still an open challenge.

In this paper, we propose a novel attack method named MetaDFME to address the distribution shift problem in DFME, aiming to stabilize the accuracy of the substitute model during the attack. Our method operates under the same practical settings as the state-of-the-art method~\cite{rosenthal2023disguide} by ensuring: \textit{black-box} (no information about the target model's architecture, parameters, and gradients), \textit{data-free} (no statistical knowledge or access to the target model's in-distribution data, and no use of public data), \textit{limited query budget}, and \textit{hard-label} (reliance solely on top-1 predictions from the target model). To achieve our objective under these settings, MetaDFME consists of a generator and an ensemble of substitute models, where the generator learns to generate data that maximizes the discrepancy among the substitute models' predictions. Meanwhile, each substitute model in the ensemble learns independently to match the target model's behavior on the generated data.

To address the distribution shift problem, we introduce a novel generator training approach based on meta-learning \cite{finn2017model} that aims to minimize distribution discrepancies across attack iterations while still maximizing the prediction disagreement among the substitute models. Specifically, our generator training has two optimization loops: an outer and an inner loop. The outer loop trains the generator to capture the consistent patterns or attributes, referred to as meta-representations, present in the synthetic data across attack iterations. In the inner loop, the generator is adapted for a few steps to produce data that incorporates the meta-representations and additional features to maximize the prediction disagreement among substitute models. Our intuition behind this approach is that the synthetic data satisfying the generator training objective among attack iterations could exhibit underlying common patterns (i.e., meta-representations). By training the generator to capture and consistently reproduce these patterns, the shift of synthetic distributions among attack iterations can be reduced.

In summary, our main contributions are listed as follows:
\begin{itemize}
    \item We propose a novel generator training approach based on meta-learning to reduce distribution shift by training the generator to capture meta-representations in synthetic data across attack iterations while allowing rapid adaptation to produce other features customized for the specific substitute models at each attack iteration.
    \item We introduce MetaDFME, a new DFME method that incorporates our novel generator training approach to effectively extract the functionality of a black-box target model with a stable performance throughout the attack.
    \item We evaluate MetaDFME on the four popular baseline image datasets, MNIST, SVHN, CIFAR-10 and CIFAR-100, and show that our method outperforms the state-of-the-art method by up to $10.3\%$ accuracy under the same attack setups. Notably, our substitute models exhibit a highly stable accuracy during the attack.
\end{itemize}

\section{Related Work}
\subsection{Data-Free Model Extraction}
Model extraction was first introduced by Tramer et al. \cite{tramer2016stealing}, where the authors demonstrated the feasibility of replicating the parameters of logistic regression models, neural networks, and decision tree classifiers. Following are two seminal works, CopycatCNN \cite{correia2018copycat} and Knockoff Net \cite{orekondy2019knockoff}, where the authors formulate the problem of model extraction as a two-stage process: data generation and substitute training. During the data generation, the authors obtain a public, unannotated dataset to query the target model and construct an annotated transfer set. This transfer set is then used to train the substitute model in the latter stage. Other works also utilize public datasets to pre-train a generative adversarial network (GAN) as a weak prior to facilitating the data generation process of model extraction attacks \cite{sanyal2022towards, lin2023quda}. These early works are foundational since they demonstrate the feasibility of extracting a black-box target model even with irrelevant public data.

In realistic attack scenarios where obtaining a public dataset for model extraction attacks is not always possible; recent research has introduced several data-free methods. The pioneers of these methods are MAZE \cite{kariyappa2021maze} and DFME-Truong\footnote{This method is named DFME by the authors, which is similar to our abbreviation of data-free model extraction in this paper. Thus, we refer to their method as DFME-Truong for better clarity.} \cite{truong2021data}. Instead of utilizing public datasets or pre-training a GAN, MAZE and DFME-Truong train a generator and a substitute model adversarially from scratch, in which the generator learns to maximize the prediction discrepancy between the substitute and the target models. Meanwhile, the substitute model learns to minimize this discrepancy. However, the generator training requires gradient back-propagation through the target model, which is impossible in black-box settings. Thus, the authors proposed using zeroth-order gradient approximation to estimate the target's gradients and train the generator.

Unfortunately, using zeroth-order gradient approximation leads to a high number of queries, making the attack costly and impractical. To overcome this, many works have proposed using the substitute model as an approximation of the target model to train the generator \cite{zhang2022ideal, beetham2023dual, rosenthal2023disguide}. Among these, Rosenthal et al. \cite{rosenthal2023disguide} proposed training an ensemble of substitute models to match the target model while the generator is trained to maximize the prediction discrepancy among substitute models. This method shows the state-of-the-art accuracy on CIFAR-10 and CIFAR-100 \cite{krizhevsky2009learning}, which are two popular baseline image datasets in the current literature. Nonetheless, the generation strategies in existing DFME methods update the generator based solely on a single, direct objective at each attack iteration without considering potential correlations with the generated data at other iterations. This could lead to distribution shifts and oscillations in the substitute model's accuracy. Meanwhile, our method learns stable meta-representations across attack iterations and dynamically adapts to the specific substitute model at each attack iteration, mitigating distribution shifts and stabilizing the substitute model's accuracy.

\subsection{Meta Learning}
Despite various successes, deep learning models have been proven to not generalize well to new tasks. To overcome this limitation, Finn et al. \cite{finn2017model} introduced a model-agnostic meta-learning approach named MAML. This approach learns an optimal initialization (i.e., meta-knowledge) for a model that can be effectively adapted to a new task with just a few samples from the task. Basically, MAML is a bi-level optimization algorithm in which the inner optimization aims to learn an optimal model for a specific task, while the outer optimization learns a meta-model (i.e., optimal initialization) that allows fast adaptation to various tasks. To solve this bi-level optimization, MAML requires high-order gradient calculation, which eventually slows down the training process. Hence, the authors introduced a first-order approximation of MAML (FOMAML) to eliminate high-order gradient calculation without sacrificing adaptation capability. Another approximation method for MAML is Reptile \cite{nichol2018first}, in which the gradients for the meta-model are obtained directly from the adapted model's weights without any gradient calculation, resulting in a simple yet effective implementation for MAML. Inspired by MAML, we formulate our generator training as a meta-learning problem to learn the meta-representations across attack iterations, which can be quickly adapted to synthesize data for the specific substitute model at each attack iteration, mitigating the distribution shift. We also adopt Reptile to accelerate our training process.

\section{Threat Model}\label{sec:threat_model}
In this section, we provide the threat model of our attack, starting with the attacker's knowledge and incentives, and ending with the attacker's actions and resources for carrying out the attack.

\textbf{Attacker's Knowledge.} 
Under a practical MLaaS context, we assume that attackers have no knowledge about the inner workings of the target model, including the model architecture, parameters, and gradients. Additionally, since the target model's development data (including all data subsets used to train, validate, or fine-tune the model) are part of the confidential MLaaS platform, and obtaining a semantically and statistically similar dataset from the internet is rarely possible \cite{truong2021data}, we also consider that no realistic data are available to attackers. Finally, we assume the attackers can only obtain categorical (hard-label) predictions from the target model. This assumption makes model extraction more realistic but challenging since less information is given to attackers, as in soft-label settings, where a minor change in the input data leads to changes in the output probabilities.

\textbf{Attacker's Incentives.}
Model extraction adversaries generally have two main incentives: \textit{theft} and \textit{reconnaissance}. In the former incentive, attackers aim to construct a substitute model whose performance on the original task is competitive with the target model. In reconnaissance incentive, attackers' ultimate goal is to launch other adversarial attacks against the target model, such as evasion attacks to fool the target model on certain samples \cite{papernot2017practical} or model inversion attacks to reconstruct the target model's training data \cite{fredrikson2015model}. To achieve this goal, the attackers attempt to construct a substitute model such that its predictions closely match the target model, including the incorrect predictions, on a wide range of inputs. \textit{In this paper, we focus on the theft incentive} and adopt the target model's test set to measure the substitute model's performance, similar to existing works \cite{rosenthal2023disguide, truong2021data, kariyappa2021maze}. 
% However, the target model's test set is only used for evaluating the substitute model during the experiments.

\textbf{Attacker's Actions.}
In black-box settings under the MLaaS context, the simplest form of interaction between the adversaries and the target is making queries. Specifically, we assume the attackers can only make queries to the target model via a prediction API and receive the corresponding hard-label predictions. Apart from queries, attackers would have no further software or hardware access to the target model to facilitate the attack with side-channel leakages.

\textbf{Attacker's Resources.}
When the target model is deployed to production, it is closely supervised by various monitoring systems to detect anomalies during the system's operations, including adversarial attempts. Also, if the attack consumes a high number of queries, it will become costly and unprofitable for attackers in the end. For these reasons, we attempt to keep the query budget low to ensure the profit and stealthiness of our model extraction attack.

\section{MetaDFME}
In this section, we provide details of MetaDFME, including its components, attack workflow, and optimization algorithm.

\subsection{Overview}

\begin{algorithm}
\SetAlgoLined
\caption{MetaDFME Algorithm}\label{alg:overview}
\SetKwInOut{Input}{Input}\SetKwInOut{Output}{Output}

\Input{target model $f^t$, generator $G_{\theta_G}$, substitute model ensemble $f^s_{\theta_S}$, memory buffer $M$}
\Output{high-accuracy substitute model ensemble $f^s_{\theta_S}$}

\For{$i$ iterations}{
    \textcolor{gray}{\emph{// Stage 1. Generator training (detailed in Section \ref{sec:g_training})}} \\
    Initialize an adapted generator $G_{\hat{\theta}_G}$ from $G_{\theta_G}$ \\
    \For{$p$ steps}{
        Adapt $G_{\hat{\theta}_G}$ for subsequent data synthesis \\
    }
    Update $G_{\theta_G}$ based on the adaptation feedback \\
    
    \textcolor{gray}{\emph{// Stage 2. Substitute training (detailed in Section \ref{sec:s_training})}} \\
    \For{$q$ steps}{
        Synthesize data with $G_{\hat{\theta}_G}$ \\
        Query $f^t$ and update $M$ \\
        Update $f^s_{\theta_S}$ on the new data pairs \\
    }
    \For{$r$ steps}{
        \If{$M$ is empty}{
            break \\
        }
        Sample previous data from $M$ \\
        Update $f^s_{\theta_S}$ on the replayed data pairs \\
    }
}
\end{algorithm}

Our attack workflow is presented in Algorithm \ref{alg:overview}. Generally, MetaDFME consists of four components: a black-box target model $f^t$, a generator $G$ parameterized by $\theta_G$, an ensemble of substitute models $f^s$ parameterized by $\theta_S$, and a memory buffer $M$. The attack procedure takes a maximum of $i$ iterations (or when the query budget is exhausted); each iteration consists of two main stages: generator and substitute training. In the former stage, the generator $G_{\theta_G}$ goes through $p$ adaptation steps, resulting in an adapted generator $G_{\hat{\theta}_G}$ for the subsequent data synthesis. This adaptation process then provides feedback for the generator $G_{\theta_G}$ to update its knowledge about the meta-representations in the synthetic data. In the substitute training stage, the adapted generator synthesizes a batch of data to query the target model. These pairs of synthetic data and hard-label predictions from the target model, in addition to some replayed data from the memory buffer $M$, are then used to train the substitute models, aiming to replicate the behaviors of the target model.

\subsection{Generator Training}\label{sec:g_training}
The main objective of the generator is to synthesize data with a balanced class distribution while promoting discrepancy in predictions among the substitute models. Also, the distributions of the synthetic data across attack iterations should have small shifts to facilitate the substitute training. To achieve this objective, we formulate our generator training as a meta-learning problem, in which the outer optimization trains the generator to capture the meta-representations in the synthetic data across attack iterations, while the inner optimization adapts the generator to customize the data with additional features according to the substitute models at the current attack iteration. Thanks to the meta-representations, the data generated by the adapted generator for the substitute training at each attack iteration have a smaller distribution shift.

From the above objective, our generator training is formulated as follows:

\begin{equation}\label{eq:meta_objective_g}
    \min_{\theta_G} \mathcal{L}_G \big( U^p_{\mathcal{L}_G}(\theta_G)\big)
\end{equation}
where $U^p_{\mathcal{L}_G}$ represents the $p$-step adaptation guided by the loss function $\mathcal{L}_G$ (will be detailed later in this section). In Equation~\ref{eq:meta_objective_g}, the outer optimization $\mathcal{L}_G(\cdot)$ acts as the meta-representation learning since it facilitates the adaptation process $U^p_{\mathcal{L}_G}$ to arrive at the solution of $\mathcal{L}_G$ minimization at each attack iteration within a few gradient descent steps.

The $p$-step adaptation process $U^p_{\mathcal{L}_G}$ takes the generator parameters $\theta_G$ and makes $p$ gradient descent steps guided by the loss function $\mathcal{L}_G$ and a learning rate $\alpha$ to produce the adapted generator parameters $\hat{\theta}_G$:

\begin{equation}\label{eq:update_g}
\begin{split}
    &\theta_G^0 = \theta_G \\
    &\theta_G^p = \theta_G^{p-1} - \alpha \nabla_{\theta_G^{p-1}} \mathcal{L}_G \big( \theta_G^{p-1} \big) \\
    &\hat{\theta}_G = \theta_G^p
\end{split}
\end{equation}

To solve the outer optimization in Equation \ref{eq:meta_objective_g}, it requires accumulating the gradients of the generator from the adaptation process, leading to high-order gradient computation. In particular, the gradient of the outer loss function with respect to the generator parameters $\theta_G$ can be computed as follows:

\begin{equation}
    \nabla_{\theta_G} \mathcal{L}_G \big( U^p_{\mathcal{L}_G}(\theta_G)\big) = \nabla_{\hat{\theta}_G}\mathcal{L}_{G} (\hat{\theta}_G) \cdot \nabla_{\theta_G} U^p_{\mathcal{L}_G}(\theta_G) 
\end{equation}
where $\hat{\theta}_G$ is obtained by using Equation \ref{eq:update_g} and fundamentally corresponding to accumulating several gradients to the initial parameters: $\hat{\theta}_G = \theta_G - \alpha\nabla_{\theta_G^1} - \alpha\nabla_{\theta_G^2} - ... - \alpha\nabla_{\theta_G^{p-1}}$. Thus, a single gradient descent of the outer optimization is computationally expensive.

To overcome this bottleneck, we treat the difference between the generator parameters $\theta_G$ and the adapted generator parameters $\hat{\theta}_G$ as a gradient and directly use this to update the generator parameters $\theta_G$ \cite{nichol2018first}. Thus, the outer update can be formulated as follows:

\begin{equation}\label{eq:g_meta_update}
    \theta_G = \theta_G + \epsilon (\hat{\theta}_G - \theta_G)
\end{equation}
where $\epsilon$ is the learning rate of the outer update.

To train the adapted generator to synthesize data with a balanced class distribution and promote discrepancy among substitute models' predictions, we construct a generator loss as a combination of discrepancy loss and diversity loss as follows:

\begin{equation}\label{eq:g_loss}
    \mathcal{L}_G = \mathcal{L}_{dis} + \lambda \mathcal{L}_{div}
\end{equation}
where $\mathcal{L}_{dis}$ and $\mathcal{L}_{div}$ stand for the discrepancy and diversity loss, respectively; $\lambda$ is a coefficient controlling the weight of the diversity loss.

The intuition behind the discrepancy loss $\mathcal{L}_{dis}$ is that, given a sample, if there is a mismatch among the substitute models, at least one of them mismatches the target model. As a result, this sample is helpful for the odd substitute models learning to match the target model. To measure this discrepancy, we calculate the standard deviation among the probability outputs of $K$ substitute models and average these standard deviations on a batch of $N$ inputs. The discrepancy loss $\mathcal{L}_{dis}$ is given below:

\begin{equation}\label{eq:l_dis}
\begin{split}
    &w_{k,n} = softmax\big(f^s_{\theta_{S_k}}(\hat{x}_n)\big) \\
    &\bar{w}_{n} = \frac{1}{K} \sum_{k=1}^K w_{k,n} \\
    &\mathcal{L}_{dis} = - \frac{1}{N} \sum_{n=1}^N \sqrt{\frac{\sum^K_{k=1} (w_{k,n} - \bar{w}_n)^2}{K}}
\end{split}
\end{equation}
in which $\hat{x}_n$ is the $n^{th}$ sample in the batch generated by the adapted generator $G_{\hat{\theta}_G}$, $w_{k,n}$ is the probability output of the $k^{th}$ substitute model on the sample $\hat{x}_n$, and $\bar{w}_n$ is the mean probability output among K substitute models on the sample $\hat{x}_n$.

To promote the balanced class distribution for the generated data, we calculate the diversity loss $\mathcal{L}_{div}$ as the information entropy of the mean probability outputs on a mini-batch of $N$ samples, as given in Equation \ref{eq:l_div}. Ideally, this diversity loss is minimized when the mean probabilities of all classes are the same and equal to $\frac{1}{C}$, where $C$ is the number of classes, resulting in a batch of high class-diversity data.

\begin{equation}\label{eq:l_div}
    \mathcal{L}_{div} = \frac{1}{N} \sum_{n=1}^N \bar{w}_n log(\bar{w}_n)
\end{equation}

\subsection{Substitute Training}\label{sec:s_training}
The main objective of the substitute models is to imitate the predictions of the target model on the data synthesized by the adapted generator. Since the data presented to the substitute models at different attack iterations still have small shifts in distribution, the substitute models should also maintain their knowledge from different synthetic data distributions during the attacks to avoid catastrophic forgetting \cite{robins1995catastrophic}. To achieve this objective, we simultaneously train the substitute models on newly generated data from the adapted generator and replayed data from the memory buffer. Notably, as our method consists of an ensemble of substitute models, we train each substitute model individually; thus, the notations related to the substitute models in this section imply an individual substitute model.

During this substitute training, the substitute models go through two sub-stages: 1) a $q$-step learning on newly generated data and 2) an $r$-step experience replay. In each step of the first sub-stage, the adapted generator $G_{\hat{\theta}_G}$ synthesizes a batch of $B$ samples $\hat{x}$ from noise vectors $z$ with its fixed parameters $\hat{\theta}_G$:
\begin{equation}
    \hat{x} = G_{\hat{\theta}_G}(z)
\end{equation}

Each synthesized sample is then queried to the target model to obtain its categorical prediction, which is used as the ground truth to optimize the substitute model with the cross-entropy loss, as defined in Equation \ref{eq:s_loss}. These data pairs are also stored in the memory buffer $M$ for later experience replay.

\begin{equation}\label{eq:s_loss}
    \mathcal{L}_S = \frac{1}{B} \sum_{n=1}^B CE\big( f^s_{\theta_S}(\hat{x}), \arg\max(f^t(\hat{x})) \big)
\end{equation}

In each experience replay step, a batch of $(x_M, y_M)$ is uniformly sampled from the memory buffer $M$ to train the substitute model with the cross-entropy loss as follows:

\begin{equation}\label{eq:s_loss_replay}
    \mathcal{L}_M = \frac{1}{N} \sum_{n=1}^N CE\big( f^s_{\theta_S}(x_M), y_M) \big)
\end{equation}

\section{Experiment}\label{sec:experiment}
In this section, we provide details of our experiments, including experimental setup and results, to demonstrate the effectiveness of MetaDFME.

\subsection{Setup}
To evaluate the effectiveness of our method, we compare MetaDFME with the most seminal work in the DFME literature, DFME-Truong \cite{truong2021data}, and the current state-of-the-art, DisGUIDE \cite{rosenthal2023disguide}. Notably, we do not compare with model extraction methods that use public data and random noise, since Truong et al. \cite{truong2021data} have systematically demonstrated DFME's superior accuracy compared to those methods. Furthermore, attacks relying on public data require the proxy dataset to be statistically and semantically similar to the target's training data (e.g., use CIFAR-10 as a proxy dataset to attack the target pre-trained on CIFAR-100), which is often infeasible and leads to significantly low accuracy if not met.

The experiments are conducted on four popular baseline image datasets, MNIST \cite{lecun1998mnist}, SVHN \cite{netzer2011reading}, CIFAR-10 \cite{krizhevsky2009learning}, and CIFAR-100 \cite{krizhevsky2009learning}. On MNIST and SVHN, the target model is LeNet \cite{lecun1998gradient} while the substitute model is LeNet-Half \cite{lecun1998gradient}. On CIFAR-10, the target and substitute models are ResNet-34-8x \cite{fang2019data} and ResNet-18-8x \cite{fang2019data}, respectively. On CIFAR-100, both the target and substitute models are ResNet-18-8x. The number of networks within the substitute model ensemble is two for all experiments. The generator is a vanilla generative network, as introduced in \cite{fang2019data}, with three convolution blocks interleaved by two $2\times$ upsampling layers. A batch-normalization layer and a $tanh$ activation are also applied at the end of the generator to normalize and scale the generated data to $[-1, 1]$, as done in \cite{rosenthal2023disguide, fang2019data, truong2021data}. Notably, despite not being stated on the paper, we recognize that the authors of DisGUIDE intentionally scale the target model's datasets to the $tanh$ domain, implicitly aligning with the attackers' generated data. Meanwhile, it is rarely possible for attackers to know the exact input range of the black-box target model. As a result, we do not scale the datasets to the $tanh$ domain in all experiments, including experiments of DisGUIDE in this work.

Regarding the target training configurations, we adopt SGD with an initial learning rate of $0.1$, a momentum of $0.9$, and a weight decay of $0.0005$ to train the target model on SVHN, CIFAR-10, and CIFAR-100 with a batch size of $256$. The number of epochs is $50$ for SVHN, and $200$ for CIFAR-10 and CIFAR-100. Meanwhile, the learning rate is $0.0001$ and the number of epochs is $20$ for MNIST. Also, the output of the target model in MetaDFME and DisGUIDE attacks is hard labels, while soft labels are used for DFME-Truong since their method requires the target's probabilities for gradient estimation.

Regarding the model extraction attacks, the generator is optimized with Adam, with an initial learning rate of $0.1$ and a weight decay of $0.0005$. The same optimization is employed for the generator adaptation process, except the adaptation learning rate is $0.0001$. On the other hand, SGD with an initial learning rate of $0.1$, a momentum of $0.9$, and a weight decay of $0.0005$ is used to train the substitute models. The batch size is set to $256$. The diversity loss' coefficient $\lambda$ is set to $0.04$ for CIFAR-100 and $0.2$ for other datasets, consistent with DisGUIDE. The query budgets on MNIST, SVHN, CIFAR-10, and CIFAR-100 for DisGUIDE and MetaDFME are $1.2M$, $2M$, $10M$ and $10M$, respectively, while those for DFME-Truong are $3M$, $4M$, $30M$, and $30M$. The memory buffer size is set to $1M$ for both MetaDFME and DisGUIDE. Also, the performance of the substitute model is evaluated every $200$ attack iterations on the held-out test set. Finally, for the number of generator and substitute training steps, we adopt the same $1:1$ setting from DisGUIDE, while our generator adaptation step is $2$.

\subsection{Results}
\subsubsection{Overall Accuracy Comparison}

\begin{table*}[h]
\centering
\renewcommand{\arraystretch}{1.5}
\resizebox{\textwidth}{!}{%
\begin{tabular}{lcccccccc}
\hline
 &
  \multicolumn{2}{c}{\textbf{MNIST}} &
  \multicolumn{2}{c}{\textbf{SVHN}} &
  \multicolumn{2}{c}{\textbf{CIFAR-10}} &
  \multicolumn{2}{c}{\textbf{CIFAR-100}} \\ \cline{2-9} 
\multirow{-2}{*}{\textbf{Method}} &
  \textbf{Query Budget} &
  \textbf{Accuracy (\%)} &
  \textbf{Query Budget} &
  \textbf{Accuracy (\%)} &
  \textbf{Query Budget} &
  \textbf{Accuracy (\%)} &
  \textbf{Query Budget} &
  \textbf{Accuracy (\%)} \\ \hline
\textbf{Target Model} & -    & 95.06          & -  & 96.48          & -   & 95.55          & -   & 77.28          \\ \hline
\textbf{DFME-Truong}         & 3M   & 55.77          & 4M & 86.25          & 30M & 59.09          & 30M & 24.26          \\
\textbf{DisGUIDE}     & 1.2M & 79.96          & 2M & 95.40          & 10M & 81.82          & 10M & 50.21          \\ \hline
\textbf{Ours}         & 1.2M & \textbf{90.26} & 2M & \textbf{95.66} & 10M & \textbf{85.24} & 10M & \textbf{56.42} \\ \hline
\end{tabular}%
}
\caption{The final accuracy of the substitute model. The best results are \textbf{bold}.}
\label{tab:overall_accuracy}
\end{table*}

In Table \ref{tab:overall_accuracy}, we compare the final accuracy of all methods on the four datasets. The final accuracy is measured on the target model's held-out test set when the query budget is exhausted. From the results, we can see that MetaDFME consistently outperforms DFME-Truong and DisGUIDE by a large margin. For instance, MetaDFME outperforms DisGUIDE by $10.3\%$ on MNIST, $3.42\%$ on CIFAR-10, and $6.21\%$ on CIFAR-100. On SVHN, DisGUIDE and our method are highly competitive, with a gap of only $0.26\%$. On the other hand, DFME-Truong performs worse than ours and DisGUIDE on all datasets while consuming significantly more queries. Notably, our accuracy on CIFAR-100 is $2.32 \times$ higher than DFME-Truong. These results demonstrate not only the superior performance of our method but also the effectiveness of our generator training approach since MetaDFME resembles DisGUIDE in the substitute training and the experimental setups while still showing an improvement of up to $10.3\%$ accuracy.

\subsubsection{Attack Stability}

\begin{table}[h]
\centering
\renewcommand{\arraystretch}{1.5}
% Consider adding resizebox back if needed, e.g.:
\resizebox{\columnwidth}{!}{%
\begin{tabular}{clcccc}
\hline
\textbf{Percentile} & \textbf{Method} & \textbf{MNIST} & \textbf{SVHN} & \textbf{CIFAR-10} & \textbf{CIFAR-100} \\
\hline
\multirow{3}{*}{$1^{st}$} & DFME-Truong     & 34.16 $\pm$ \textbf{13.09} & 17.20 $\pm$ \textbf{04.72} & 22.94 $\pm$ \textbf{01.98} & 08.59 $\pm$ \textbf{01.38} \\
                          & DisGUIDE & 53.96 $\pm$ 21.23          & 61.71 $\pm$ 21.82          & 41.90 $\pm$ 03.48          & 12.94 $\pm$ 01.71          \\
                          \cline{2-6}
                          & \textbf{Ours}     & \textbf{74.31} $\pm$ 22.68 & \textbf{70.07} $\pm$ 21.74 & \textbf{54.39} $\pm$ 05.50 & \textbf{21.76} $\pm$ 03.27 \\
\hline
\multirow{3}{*}{$2^{nd}$} & DFME-Truong     & 58.89 $\pm$ 03.07          & 70.12 $\pm$ 10.33          & 46.61 $\pm$ 01.33          & 18.62 $\pm$ \textbf{00.55} \\
                          & DisGUIDE & 74.88 $\pm$ 02.74          & 92.77 $\pm$ 00.89          & 61.20 $\pm$ 02.01          & 27.39 $\pm$ 01.63          \\
                          \cline{2-6}
                          & \textbf{Ours}     & \textbf{91.25 $\pm$ 01.05} & \textbf{95.01 $\pm$ 00.16} & \textbf{76.37 $\pm$ 00.46} & \textbf{45.35} $\pm$ 00.63 \\
\hline
\multirow{3}{*}{$3^{rd}$} & DFME-Truong     & 58.92 $\pm$ 00.51          & 83.32 $\pm$ 00.89          & 54.33 $\pm$ 00.51          & 22.23 $\pm$ \textbf{00.16} \\
                          & DisGUIDE & 77.88 $\pm$ 02.43          & 94.41 $\pm$ 00.25          & 73.40 $\pm$ 00.80          & 38.21 $\pm$ 00.63          \\
                          \cline{2-6}
                          & \textbf{Ours}     & \textbf{90.72 $\pm$ 00.35} & \textbf{95.30 $\pm$ 00.14} & \textbf{81.52 $\pm$ 00.25} & \textbf{53.46} $\pm$ 00.29 \\
\hline
\multirow{3}{*}{$4^{th}$} & DFME-Truong     & 56.51 $\pm$ 00.49          & 85.37 $\pm$ 00.55          & 56.97 $\pm$ 00.46          & 23.35 $\pm$ \textbf{00.18} \\
                          & DisGUIDE & 80.85 $\pm$ 01.90          & 95.22 $\pm$ 00.24          & 80.94 $\pm$ 00.73          & 46.47 $\pm$ 00.78          \\
                          \cline{2-6}
                          & \textbf{Ours}     & \textbf{90.41 $\pm$ 00.17} & \textbf{95.53 $\pm$ 00.08} & \textbf{82.95 $\pm$ 00.35} & \textbf{55.54 $\pm$ 00.18} \\
\hline
\end{tabular}%
} % End resizebox if used
\caption{The mean accuracy (in \%) $\pm$ the $95\%$ confidence interval (in \%) within the four query budget percentiles of all attack methods. The higher mean accuracy and the lower confidence interval are better. The best results are \textbf{bold}.}
\label{tab:accuracy_percentile}
\end{table}

Subsequently, we analyze the stability in the substitute model's accuracy of all methods throughout the attack. This is particularly important in DFME because attackers would not have the target model's in-distribution data to evaluate the substitute model's performance during the attack; therefore, the attack's effectiveness entirely relies on the attack's random termination. Thus, a practical DFME attack should maintain the stable performance of the substitute model throughout the attack process; otherwise, the substitute model's final performance would be sub-optimal.

To demonstrate the stability in the substitute model's accuracy of our method, we measure the mean accuracy and the $95\%$ confidence interval within the four percentiles of the query budget: $0-25\%$, $25 - 50\%$, $50 - 75\%$, and $75 - 100\%$. This setup aims to estimate the accuracy of the substitute model that the attacker can expect when randomly terminating the attack within these percentiles. The results are shown in Table \ref{tab:accuracy_percentile}.

Generally, although all methods improve the substitute model's accuracy and stability throughout the attack, a better result is shown in MetaDFME. On CIFAR-10, our method starts with $54.39\%$ mean accuracy in the first query budget percentile, while DisGUIDE only achieves $41.90\%$. Notably, our confidence interval is $\pm 5.50\%$, which is higher than DisGUIDE with $\pm 03.48\%$, demonstrating more fluctuation. One possible reason for this fluctuation is that our generator \textit{"is seeking"} the meta-representations in the synthetic data during this early attack period; therefore, the generated data might exhibit some extent of distribution shifts, causing challenges for the substitute model. Nonetheless, the performance of the substitute model immediately stabilizes from the next query budget percentile with a high mean accuracy and a tight confidence interval, drastically better than DisGUIDE. For instance, MetaDFME achieves $76.37\% \pm 00.46\%$ and $81.52\% \pm 00.25\%$ on the next two percentiles, while DisGUIDE only achieves $61.20\% \pm 02.01\%$ and $73.40\% \pm 00.80\%$. The same observations hold for MNIST, SVHN, and CIFAR-100. Notably, DFME-Truong exhibits the tightest confidence interval on CIFAR-100, which is attributed to the slow learning curve of this method. In summary, with our MetaDFME method, attackers can obtain a substitute model with a higher accuracy and higher confidence than the current state-of-the-art.

\subsubsection{Query Efficiency}

\begin{figure}[h]
    \centering
    \includegraphics[width=\columnwidth]{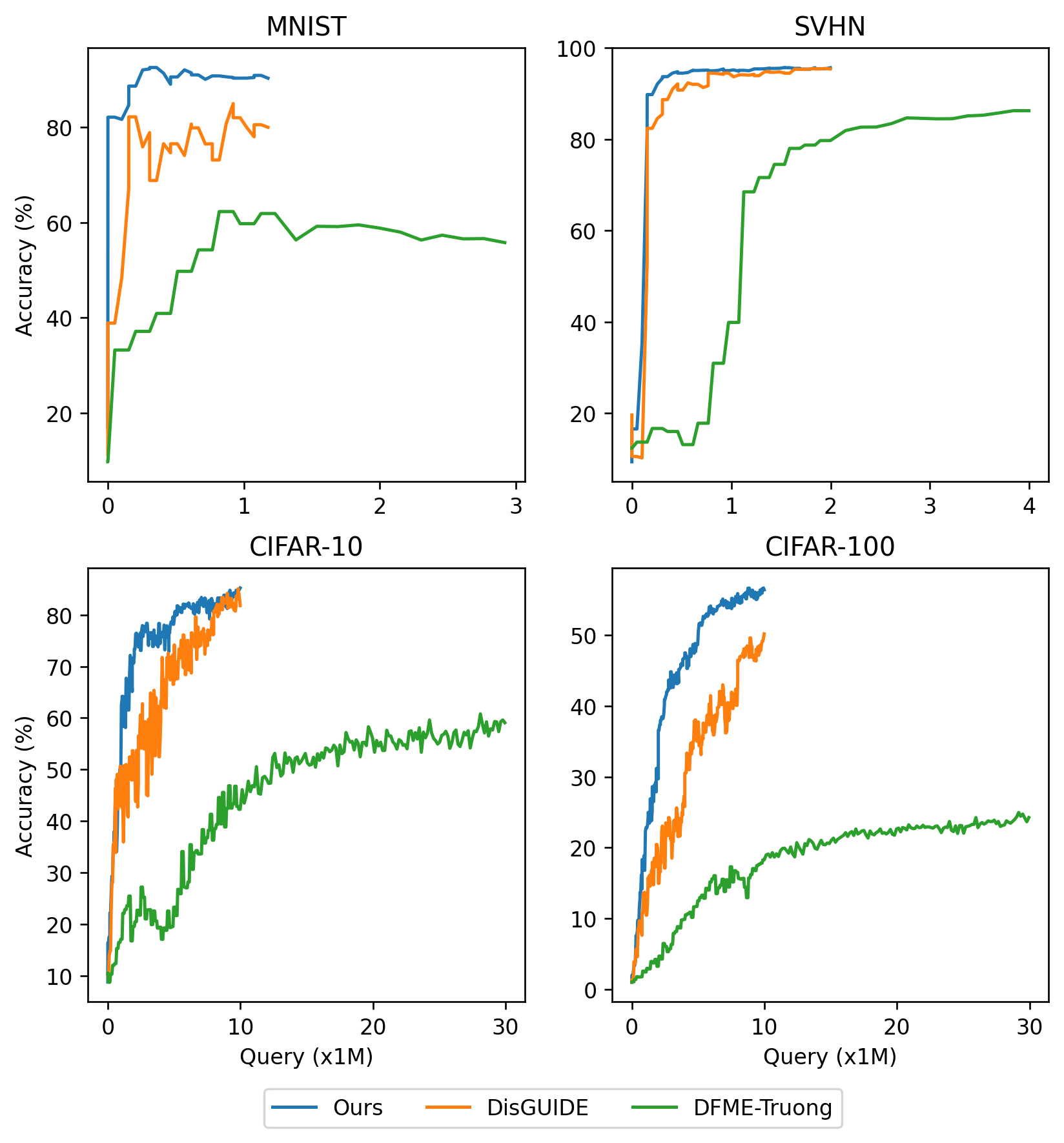}
    \caption{The accuracy evolution of the substitute model during the attack.}
    \label{fig:accuracy_evolution}
\end{figure}

As part of the attacker's resource, we also analyze the query efficiency of MetaDFME, DisGUIDE, and DFME-Truong. Figure \ref{fig:accuracy_evolution} visualizes the accuracy of the substitute model on the target model's held-out test set with respect to the number of queries during the attack. Overall, our method demonstrates a better accuracy evolution on all datasets. For example, on CIFAR-10, our substitute model quickly achieves $75\%$ accuracy at around $2M$ queries and then gradually improves to the end of the attack. On the other hand, DisGUIDE's substitute model learns more slowly while exhibiting significant oscillation throughout the attack. The same pattern is observed on other datasets, where MetaDFME witnesses a better accuracy improvement with less oscillation than DisGUIDE. Notably, DFME-Truong exhibits the slowest learning evolution. For instance, the accuracy of DFME-Truong at $30M$ queries on CIFAR-100 is less than half of DisGUIDE and our method at $10M$ queries.

\section{Ablation Studies}
In this section, we investigate the effectiveness of our method regarding different generator training setups.

\subsection{Adaptation Steps}

\begin{table}[]
\centering
\renewcommand{\arraystretch}{1.5}
\resizebox{0.7\columnwidth}{!}{%
\begin{tabular}{lccc}
\hline
                               & \textbf{1 step} & \textbf{2 steps} & \textbf{5 steps} \\ \hline
\textbf{Accuracy (\%)}         & 81.00           & 85.24            & 83.22            \\
\textbf{Attack Duration (hrs)} & 3.7             & 4.4              & 6.5              \\ \hline
\end{tabular}%
}
\caption{The impact of generator adaptation steps on MetaDFME on CIFAR-10.}
\label{tab:generator_adaptation}
\end{table}

\begin{figure}
    \centering
    \includegraphics[width=0.8\columnwidth]{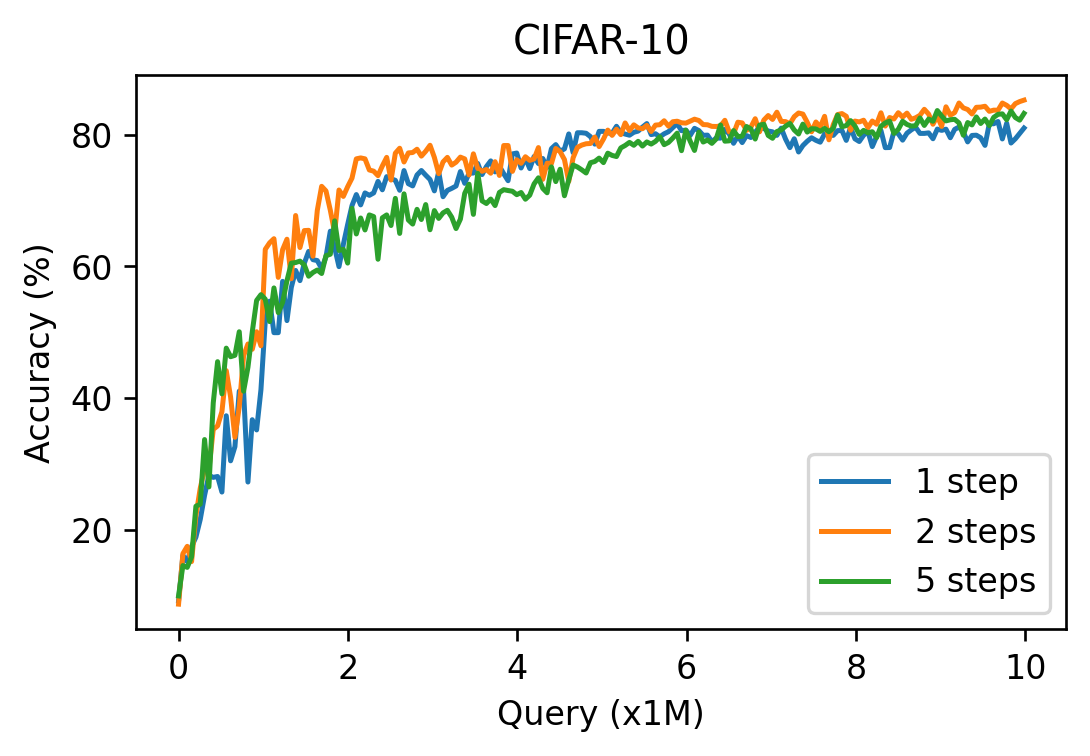}
    \caption{The impact of generator adaptation steps on the accuracy evolution of MetaDFME on CIFAR-10.}
    \label{fig:generator_adaptation}
\end{figure}

Firstly, we investigate the impact of the generator adaptation steps on the accuracy of the substitute model and the attack duration. We conduct these experiments on CIFAR-10 with the same setups as in the previous section and present the results in Table \ref{tab:generator_adaptation}. Apparently, the $2$-step adaptation setup exhibits the best trade-off between the substitute model's accuracy and attack duration. Interestingly, adapting the generator with $5$ steps results in a worse accuracy while consuming $1.5 \times$ time. In Figure \ref{fig:generator_adaptation}, we also illustrate the accuracy of the substitute model throughout the attack under different generator adaptation steps, which shows minimal difference among numbers of steps. As a result, we select $2$ as the number of generator adaptation steps in our previous experiments.

\subsection{Learning Rates}

\begin{figure}
    \centering
    \includegraphics[width=0.7\columnwidth]{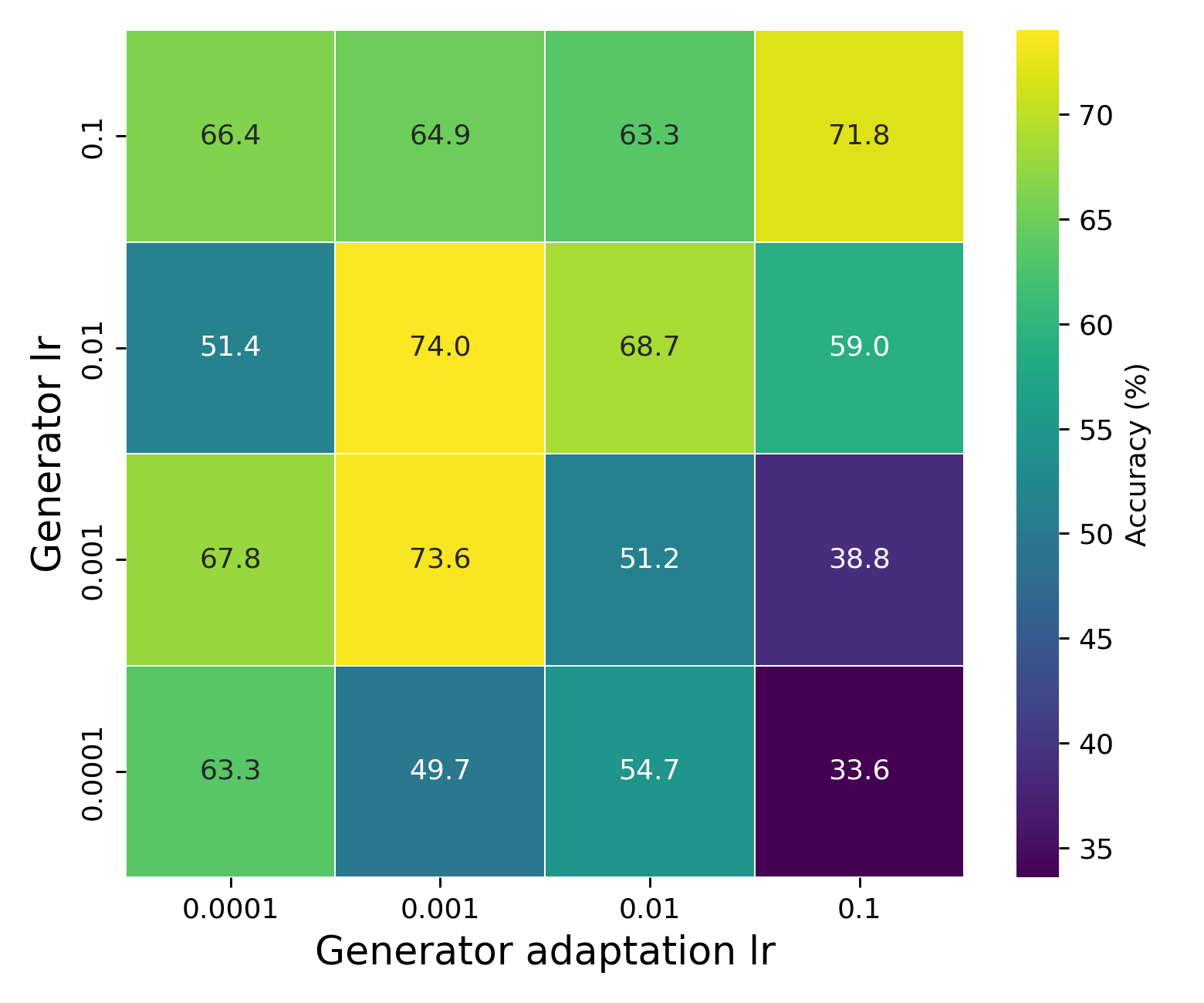}
    \caption{The impact of learning rates in generator training and generator adaptation on MetaDFME on CIFAR-10.}
    \label{fig:generator_lr}
\end{figure}

Finally, we investigate the impact of different learning rates in generator training and generator adaptation on the substitute model's accuracy. We conduct these experiments on CIFAR-10 with the same setups as in the previous section, except the query budget is only $2M$ to save time. Based on the results in Figure \ref{fig:generator_lr}, there appears a pattern that the learning rate of the generator adaptation process should not be higher than the learning rate of the meta-representation learning process. For instance, the accuracy of the substitute model is the worst when we adapt the generator with a learning rate of $0.1$ while learning the meta-representation with a learning rate of $0.0001$. This is reasonable since we want the generator to quickly capture the meta-representations with a higher learning rate, while slowly adapting the generator to fine-tune the synthetic data with a lower learning rate to facilitate the substitute training.

\section{Discussion}
This section discusses some limitations, potential improvements of our work, and some defense approaches against DFME attacks.

\subsection{Limitations and Future Work}
Despite the promising results, our work has some limitations that require further investigation to better understand and improve our attack's effectiveness. Firstly, although there are other model architectures and popular image datasets (such as ImageNet), we could only demonstrate the effectiveness of our method on the same setup as the current state-of-the-art due to our limited time. Therefore, experiments on other models and datasets are useful to thoroughly consolidate the effectiveness of our method. Secondly, there is still a performance gap between our substitute model and the target model. Although we might achieve a comparable accuracy with the target model with more queries, this makes the attack unprofitable and unstealthy. As a result, other techniques (such as semi-supervised learning or self-supervised learning) to boost the performance of the substitute model without consuming the query budget are open for future work. Finally, although our generator training approach to alleviating the effect of distribution shift by learning common features (i.e., meta-representations) is intuitive, more investigation is required to verify whether all datasets have such common features. If there are none, it necessitates other approaches and methods to handle this case. Also, more empirical experiments are necessary to understand what exactly these meta-representations are.

\subsection{Defense against DFME}
As DFME becomes more effective and practical, it urges defense methods to safeguard ML models in MLaaS, which eventually protects model owners from financial compromise and legitimate users from security and privacy violations. There are two main approaches to defend a model from DFME attacks: \textit{detection} and \textit{prevention}. The detection approach aims to identify a model theft by embedding a watermark into a protected model \cite{jia2021entangled}. If an adversary extracts this model, the watermark is transferred to the substitute model. This enables the model owners to identify the thieves by verifying the presence of the watermark in the substitute model. The main limitation of this approach is that the watermark would be useless if the model owners have no white-box access to the adversary's substitute model to verify the watermark. Also, this approach can not prevent the adversary from using the substitute model to launch other attacks against the target model. Alternatively, rate limits can also be applied to prevent attackers from making a large number of queries during DFME attacks. However, we believe that attackers can easily bypass rate limits by distributing the queries to multiple MLaaS accounts.

In contrast, the prevention approach aims to prevent the adversary from learning a high-quality substitute model in the first place. An example of this approach is MisGUIDE \cite{gurve2024misguide}, which consists of two stages to prevent a DFME attack. Firstly, the prediction APIs proactively monitor the distributions of the queried data to identify model extraction attempts since synthetic data from DFME tends to be noisier compared to the target model's in-distribution data. If an extraction attempt is identified, the prediction APIs intentionally return specially crafted predictions to mislead the substitute training of the attackers. Consequently, the substitute model from the detected DFME attack has low quality, making it useless for attackers to obtain any financial benefits or launch other attacks.

By introducing MetaDFME, we aim to facilitate AI security experts in analyzing and understanding the capabilities of DFME attacks, shedding light on safeguarding MLaaS models in real-world applications. For instance, by demonstrating that our method can obtain a decent substitute model with $2M$ queries, defenders can configure the rate limit of their applications accordingly. Furthermore, the meta-representations of our method can also be utilized by defenders to further understand the generated data distribution of DFME attacks, which could help in devising effective DFME detection methods. 

\section{Conclusion}
In this paper, we introduce MetaDFME, a novel DFME method, to address the oscillating accuracy of the substitute model during the attack due to distribution shift. Our method consists of a generator and an ensemble of substitute models. The substitute model ensemble individually learns to match the target model, while the generator learns to synthesize class-balancing data that maximizes the prediction disagreement among the substitute models. To reduce the effect of distribution shift, we train the generator to learn the meta-representations in the synthetic data across attack iterations. By capturing the meta-representations, our generator can be adapted with only a few steps to produce other features that facilitate the substitute model training at each attack iteration. Experimental results on MNIST, SVHN, CIFAR-10, and CIFAR-100 show that MetaDFME outperforms the current state-of-the-art while exhibiting a more stable performance during the attack. We also give insights into the adaptation steps and learning rates in our generator training method. From this work, we encourage future research to focus on addressing this critical problem of distribution shift in DFME to improve the attack's effectiveness and stability, eventually facilitating the development of defensive methods to protect ML models in the wild. 

%%%%%%%%%%%%%%%%%%%%%%%%%%%%%%%%%%%%%%%%%%%%%%%%%%%%%%%%%%%%%%%%%%%%%%%%

%%% Use this environment to include acknowledgements (optional).
%%% This will be omitted in doubleblind mode.

\begin{ack}
This research is funded by the University College Dublin, School of Computer Science, Ireland.
\end{ack}

%%%%%%%%%%%%%%%%%%%%%%%%%%%%%%%%%%%%%%%%%%%%%%%%%%%%%%%%%%%%%%%%%%%%%%%%

%%% Use this command to include your bibliography file.

\bibliography{mybibfile}

\end{document}